\documentclass[
	a4paper, % Paper size, use either a4paper or letterpaper
	10pt, % Default font size, can also use 11pt or 12pt, although this is not recommended
	unnumberedsections, % Comment to enable section numbering
	twoside, % Two side traditional mode where headers and footers change between odd and even pages, comment this option to make them fixed
]{t0003}

\usepackage{float}
\usepackage[title]{appendix}
\usepackage{xcolor}

\newcommand{\blue}[1]{\textcolor{blue}{#1}}
\newcommand{\gray}[1]{\textcolor{gray}{#1}}
\newcommand{\red}[1]{\textcolor{red}{#1}}

\runninghead{Natural Language Models for Data Visualization}
\title{Natural Language Models for Data Visualization\\ Utilizing nvBench Dataset} % Article title

\author{
	Shuo Wang and Carlos Crespo-Quinones
}

\date{\footnotesize DATASCI 266: Natural Language Processing \\ UC Berkeley School of Information \\ \{shuo.wang, carlos.d.crespo\}@ischool.berkeley.edu}

\renewcommand{\maketitlehookd}{%
	\begin{abstract}
		\noindent Translation of natural language into syntactically correct commands for data visualization is an important application of natural language models and could be leveraged to many different tasks. A closely related effort is the task of translating natural languages into SQL queries, which in turn could be translated into visualization with additional information from the natural language query supplied\cite{Zhong:2017qr}. Contributing to the progress in this area of research, we built natural language translation models to construct simplified versions of data and visualization queries in a language called Vega Zero first proposed by Luo, Yuyu, et al\cite{Luo:2021qr}. In this paper, we explore the design and performance of these sequence to sequence transformer based machine learning model architectures using large language models such as BERT as encoders to predict visualization commands from natural language queries, as well as apply available T5 sequence to sequence models to the problem for comparison.
	\end{abstract}
}

\begin{document}
\maketitle % Output the title section

%----------------------------------------------------------------------------------------
%	ARTICLE CONTENTS
%----------------------------------------------------------------------------------------

\section{Introduction}

Data visualization is an integral part of data analysis and reporting, highly skilled professionals spend a significant portion of their working hours constructing data queries and turning them into charts and graphs. The potential application of machine learning models to understand the underlying data and translate natural language queries into data visualization would greatly enhance the efficiency and productivity of many data analysis tasks. Not only would such tools enhance our existing capability to understand and communicate with our data, it could also help us uncover previously hidden information.

Some immediate applications include: creating bar, line and scatter plots of SQL tables, transform and group data within a table and display the aggregated data, as well as joining and combining tables of data to extract information and display them visually.

Previous research in this area has included explorations of transforming natural language into SQL commands\cite{Zhong:2017qr, Yu:2019qr} with deep learning models and natural language to visualization interfaces\cite{Cox:2001qr}. NvBench is a new dataset created to facilitate this type of research to further integrate the process of natural language to data query and data query to visualizations\cite{Luo:2021qr} . Leveraging this data set and the ncNet transformer model architecture\cite{Luo:2022qr}, we further explore the various modeling possibilty in our work.

In this paper, we have used the nvBench as our train, validation and test dataset to create BERT encoder based multi-head attention transformer models and compare the performance of the new models with the performance of the ncNet model (also a sequence to sequence transformer model). Our goal is to explore the possibility of a generalized natural language to visualization process where the ability of the system to handle natural language inputs is not limited by the input training data. We have provided below the main results of our research and detailed description and analysis of the input data, model architecture and model performance.
From our reseach, the BERT encoder based sequence to sequence transformer model was able to achieve an overall accuracy of 79\%, confirming our hypothesis of the possibility to leverage transferred learning of BERT models to data visualization tasks. Applying a pre-trained CodeT5 model realizes even more impressive results, achieving an overall accuracy of 97\%.

\section{Background}

Natural language to visualization research has been conducted for decades, some of these efforts include early approaches to manually program rule-based logic for the handling of anticipated user inputs in multi-model systems\cite{Cox:2001qr} and more recent attempts to incorporate machine learning and optimization techniques\cite{Aurisano:2016qr}. However these systems are limited in functionality by the ability of the system designers to anticipate the possible universe of inputs from users. But what if the user uses a word that the system has never seen? What if the natural language input was written in an idiomatic way that was not present in the training data? These limitations could potentially be addressed with the recent advances in large language models such as BERT and GPT through transfer learning.

Recent research exists for applying BERT model to the task of data visualization\cite{Can:2021qr} by converting natural language into vector representation embeddings and use these embeddings to predict various ingredients needed for the visualization of tabular data such as chart type, data columns and aggregation type. However, the structured approach to the problem inherently limits the expressive power of the system to generate complex visualizations.

A new dataset, nvBench\cite{Luo:2021qr}, was made publicly available for research purposes in 2021. The dataset contains pairs of natural language queries and data visualization commands in vega zero syntax\cite{Luo:2022qr}. A self-contained sequence-to-sequence model, ncNet\cite{Luo:2022qr}, created for and trained on the dataset was also made available. We base our research on this dataset and the ncNet transformer model, modifying the architecture to use BERT and other language models to train and test generalizable natural language to data visualization models, eventually applying the pre-trained T5 model to the problem and achieving superior results.

\section{Methods}

\subsection{Data}

\subsubsection{nvBench} The data set contains 7247 visualization queries (labels). Each query contains a visualization type and data retrieval command meant to be run against an underlying database with multiple tables, which contain the actual data to be displayed. Below is an example of a visualization query:

\begin{quote}
Visualize \blue{BAR} \gray{SELECT JOB\_ID , SUM(MANAGER\_ID) FROM employees WHERE salary BETWEEN 8000 AND 12000 AND commission\_pct != "null" OR department\_id != 40 GROUP BY JOB\_ID ORDER BY JOB\_ID ASC}
\end{quote}

\begin{figure}
	\includegraphics[width=\linewidth]{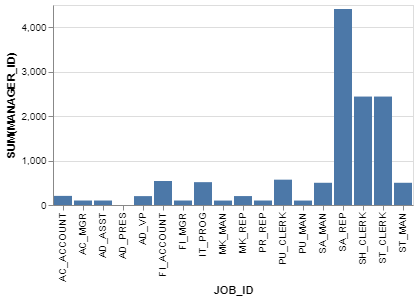}
	\caption{Rendering: Visualize \blue{BAR} \gray{SELECT JOB\_ID , SUM(MANAGER\_ID) FROM employees WHERE salary BETWEEN 8000 AND 12000 AND commission\_pct != "null" OR department\_id != 40 GROUP BY JOB\_ID ORDER BY JOB\_ID ASC}.}
	\label{fig:visualization}
\end{figure}

In this example, the chart type is ``BAR'', followed by the SQL statement for data retrieval. Figure \ref{fig:visualization} shows the rendering result of the query.

The example visualization query is mapped to several natural language queries, from which we train our models:

\begin{quote}
1. For those employees whose salary is in the range of 8000 and 12000 and commission is not null or department number does not equal to 40, show me about the distribution of job\_id and the sum of manager\_id , and group by attribute job\_id in a bar chart, I want to sort in ascending by the bar.

2. For those employees whose salary is in the range of 8000 and 12000 and commission is not null or department number does not equal to 40, a bar chart shows the distribution of job\_id and the sum of manager\_id , and group by attribute job\_id, and could you order x axis in asc order?
\end{quote}

In total, there are 25762 natural language queries, averaging 3.55 natural language queries for every visualization query.

\subsubsection{Augmented Data} Due to the complexity of SQL syntax, a simplified version of the visualization query is used in place of the original query, called vega zero\cite{Luo:2022qr}:

\begin{quote}
mark \blue{bar} data \blue{employees} encoding x \blue{hire\_date} y aggregate \blue{count hire\_date} transform filter \blue{salary between 8000 and 12000 and commission\_pct != "null" or department\_id != 40} bin x by \blue{month}	
\end{quote}

Where the visualization query is flattened into a sentence with special markers designating locations of information, please refer to \cite{Luo:2022qr} for complete syntax. In our research, we follow the same syntax rules.

Since this syntax only supports single table queries, it is not possible to predict queries where the joining of tables is necessary, so we only train and test our models on a subset of nvBench dataset, with 2988 visualization queries for training, 186 visualization queries for validation and 625 visualization queries for testing.

The natural language queries are also augmented to include the template of vega zero queries:

\begin{quote}
\blue{<N>} For those employees whose salary is in the range of 8000 and 12000 and commission is not null or department number does not equal to 40 , draw a line chart about the change of department\_id over hire\_date , display by the X from low to high . \blue{</N>} \blue{<C>} mark \blue{[T]} data employees encoding x \blue{[X]} y aggregate \blue{[AggFunction] [Y]} color \blue{[Z]} transform filter \blue{[F]} group \blue{[G]} bin \blue{[B]} sort \blue{[S]} topk \blue{[K] </C> <D>} employees \blue{<COL>} hire\_date salary department\_id last\_name first\_name \blue{</COL> <VAL>} Bull Lex Seo Bell Chen Lee Gee Banda King Baer Fay \blue{</VAL> </D>}
\end{quote}

The natural language query is enclosed in <N> </N> markers, the vega zero template is enclosed in <C> </D> markers and the locations of vega zero query fields are marked by special tokens: [T], [X], [Y], [Z], [F], [G], [B], [S] and [K]. These are the syntax used by the original ncNet sequence to sequence transformer models and we follow the same rules for our models as well.
Finally, every natural language and visualization query pair in the dataset is augmented with two versions, one with the chart type kept as placeholder, [T], and another with the chart type filled in (bar, line, etc). The motivation is to train the model to predict the visualization query whether the chart type is explicitly specified or not.

\subsubsection{Data Loader} The augmented data were provided in the Github repo for the ncNet paper\cite{Luo:2022qr} , which we use unaltered. However, the original data loader no longer works the latest version of Pytorch, we rewrote the data loader to be used for training and testing the ncNet model and our own models.

\subsection{Model}

\subsubsection{ncNet} The ncNet\cite{Luo:2022qr} model is a sequence to sequence multi-head transformer model, where the encoder transforms the natural language queries into embeddings combining position and token type information, and apply multiple layers of multi-head attention transformers to incorporate contextual information into the encoded embeddings. Shown in Figure \ref{fig:ncnet}, The decoder then takes in the encoded embeddings and the embeddings of the previously predicted tokens to further transform the predicted tokens with attention to the encoded embeddings. We use the ncNet model as our baseline, and replace the encoder with various different architectures.

One important fact to note is that the ncNet model uses all words occurring in the training, testing and validation dataset as vocabulary. Therefore limiting the range of possibilities of natural language query inputs.

We also removed the attention mask (referred to as attention forcing in Luo 2022\cite{Luo:2022qr}) used in ncNet to create another baseline for our experiments, since attention masks are not applicable to our models.

\subsubsection{BERT Model} We then proceeded to replace the ncNet encoder with the BERT model encoder, as shown in Figure \ref{fig:ncbert}. Because the BERT model was pre-trained with a much larger vocabulary, the hope was that this knowledge would transfer to the new model, ncBERT, once we fine-tuned it. We trained an additional BERT encoder model with convolutional layers added to the BERT embeddings, in order to distill relevant information from BERT model embeddings before decoding, as the BERT embedding has a much larger dimension that may contain information irrelevant to the data visualization task. The convolutional layers incorporated model is shown in Figure \ref{fig:ncbertcnn}.

\subsubsection{Combining ncNet and BERT Model} Furthermore, we created an encoder combining both the ncNet and BERT encoders, we achieved this by reshaping the embeddings from BERT to the same dimension as ncNet encoder and concatenating them together. The architecture of the model could be found in Figure \ref{fig:ncnetbert}. The motivation of creating this model was to analyze the effect of BERT embedding on the baseline model and explore the potential of enhancing the original encoder without completely replacing it. However, the implication of this approach was that words that did not exist in the original ncNet encoder still could not be used, further research needs to be done to remove this limitation.

\subsubsection{CodeT5} Finally, a series of experiments were conducted by fine-tuning two different families of the pre-trained T5 model on the nvBench dataset. The first set of models was conducted using variations of the original T5 model \cite{Raffel:2020qr} by Raffel et al. and the second set of experiments was based on the Code-T5 model family \cite{Yue:2021qr2} by Wang et al., which is a variation of the T5 model pre-trained for code specific tasks. The experiments included using models of different parameter sizes and modifying the input by including or not, the query templates along with the input NL request. The results are very encouraging, where we achieved 98\% prediction accuracy and 88\% guided search accuracy using the “large” (700M parameter) version of the CodeT5 model and using the same input sequence as the other models which included both the template and the NL request. Figure \ref{fig:codetf} shows the architecture of the model.

\subsection{Evaluation}

The models are trained and tested with training dataset of size 25238, validation dataset of size 1430 and test dataset of size 4920. Each model is run over at least 5 epochs with a learning rate of 0.0005 and the resulting losses are recorded.
Then the accuracy of the models are evaluated by running predictions with the models over the test dataset, with the natural language query tokens and the first n-1 tokens of the label as input (the label tokens are masked so that successive predictions of the label are not affected by future label tokens), counting the total number of correct predictions and dividing it by the total number of predictions.

Finally, we evaluate the models one more time with a guide search algorithm for prediction. The algorithm starts the prediction with the start of the sentence token, then repeatedly predicts the next label token with guidance until the end of the sentence token is reached. This evaluation provides a more realistic measure of the model's usability. The accuracy from this evaluation is defined as the total number of complete label matches over the total number of test labels.

\section{Results and Discussion}

\subsection{Accuracy}

After running 5 epochs with learning rate 0.0005 and keeping the model version with the lowest validation loss, the test accuracy of the models are reported in Table \ref{tab:accuracy}. The accuracy of a model is defined as the probability of predicting the next word correctly given part of the label query.

\begin{table} % Single column table
	\caption{Model test accuracy.}
	\centering
	\begin{tabular}{lccc}
		\toprule
		Model & Query & Query+Chart & Overall \\
		\midrule
		ncNet & 95\% & 96\% & 96\% \\
		ncNet w/o AF & 95\% & 96\% & 96\% \\
		\hline
		nvBERT & 79\% & 79\% & 79\% \\
		nvBERT\_CNN & 79\% & 79\% & 79\% \\
		nvncNetBERT & 89\% & 89\% & 89\% \\
		\hline
		codeT5 w/o TPT & 97\% & n/a & 97\% \\
		codeT5 & 97\% & 97\% & 97\% \\
		codeT5 Large & 98\% & 98\% & 98\% \\
		\bottomrule
	\end{tabular}
	\label{tab:accuracy}
\end{table}

The table shows three columns, the accuracy of the models on queries where chart type is not specified (Query), the accuracy of the models on queries where chart type is specified (Query+Chart) and the overall accuracy (Overall).

As we can see from the results, the original ncNet models produced test accuracies in excess of 95\%. While the BERT based transformer models were able to achieve close to 80\% accuracy. This is attributable to the fact that the BERT model contains a much larger vocabulary than the ncNet model trained only on the vocabulary of tokens present in the dataset.

Furthermore, the removal of attention forcing in the ncNet model does not appear to affect the test accuracy, suggesting that its absence in nvBERT models had no bearing on the accuracy results of the nvBERT models. The accuracy of nvBERT models with and without convolutional layers does not appear to affect the accuracy at all, suggesting minimal impact of the convolutional layers.
Finally the combined ncNet and BERT encoder produced results that improved upon the nvBERT only models, suggesting that were a model able to combine the ncNET and BERT encoder while remaining generalizable, the results would improve along with the benefits of transfer learning.

CodeT5 models achieved superior results across the board, surpassing both the original ncNet model as well as the BERT encoder integrated models. What is more exciting is the fact that, even without templates as input, codeT5 models are still able to achieve comparable results.

Besides the overall accuracy difference, the general trend is that the accuracy for ``query only tests are slightly'' lower than accuracy for ``query and chart type'' tests. This is expected since the query only tests require the model to predict an additional field. However, given that the accuracy differences are small, the chart type prediction is generally accurate.

\subsection{Guided Search Accuracy}

We next look at the accuracy of predicting the entire label query. The algorithm of guided search, as used by the original ncNet model, proceeds as follows: starting with the start of sentence token for the label, successively predict the top five candidates. Within the top five candidates, the top candidate is chosen unless it does not belong to the possible words for the current position being predicted, in which case the rest of the candidates are iterated over to find a suitable choice. The prediction is counted as correct if the resulting sentence matches the label sentence completely. Table \ref{tab:guidedaccuracy} shows the results for each model.

\begin{table} % Single column table
	\caption{Model test accuracy with guided search.}
	\centering
	\begin{tabular}{lccc}
		\toprule
		Model & Query & Query+Chart & Overall \\
		\midrule
		ncNet (baseline) & 62\% &72\% & 67\% \\
		ncNet w/o AF & 65\% & 75\% & 70\% \\
		\hline
		nvBERT & 3\% & 13\% & 8\% \\
		nvBERT\_CNN & 2\% & 12\% & 7\% \\
		nvncNetBERT & 21\% & 33\% & 27\% \\
		\hline
		codeT5 w/o TPT & 83\% & n/a & 83\% \\
		codeT5 & 86\% & 87\% & 86\% \\
		codeT5 Large & 89\% & 88\% & 88\% \\
		\bottomrule
	\end{tabular}
	\label{tab:guidedaccuracy}
\end{table}

From looking at the results, it's immediately clear that the lower accuracy in the BERT based models are amplified through the guided search process, resulting in a low overall accuracy of 7.8\%. This effect also shows up in the difference between query only and query with chart type accuracy, where the small difference in Table \ref{tab:accuracy} becomes much larger in Table \ref{tab:guidedaccuracy}.

The codeT5 models again show great results. Furthermore, the results for query only and query plus template again show comparable results, demonstrating the power of the codeT5 model decoder.

\subsection{Sample Analysis}

\subsubsection{nvBERT} Given the large discrepancy between the overall accuracy and the guided search accuracy due to the flaws of the respective metrics, we need a better understanding of the true performance of the new models with BERT encoders. One of the things to observe is that in the predicted query, many of the words are part of the template and rarely changes. These words need to be separated out because the correction prediction of them mostly requires the model to simply be position aware. As an example:

\begin{quote}
\gray{Query}: <N> Plot how many booking start date by grouped by booking start date as a bar graph , order Y in ascending order . </N> <C> \blue{mark} [T] \blue{data} apartment\_bookings \blue{ encoding} \red{x} [X] \red{y} \blue{aggregate} [AggFunction] [Y] \red{color} [Z] \blue{transform} \red{filter} [F] \red{group} [G] \red{bin} [B] \red{sort} [S] \red{topk} [K] </C> <D> apartment\_bookings <COL> booking\_start\_date </COL> <VAL> </VAL> </D>
\end{quote}

\begin{quote}
\gray{Label}: \blue{mark} bar \blue{data} apartment\_bookings \blue{encoding} \red{x} booking\_start\_date \red{y} \blue{aggregate} count booking\_start\_date \blue{transform} \red{sort} \red{y} asc \red{bin} \red{x} by weekday <eos>
\end{quote}

Words such as ``mark'', ``data'', ``encoding'', ``aggregate'' and ``transform'' always appear in the label and in fixed positions, the model is able to predict them almost perfectly. Words that are also part of the template such as ``x'', ``y'', ``color'', ``filter'', ``group'', ``bin'', ``sort'' and ``topk'' may not occur in the label or may occur more than once, are harder to predict.

Table \ref{tab:count} shows the count of words that are part of template vs non-template, on average, there are more template words than non-template words, which partially account for why word prediction accuracy is much higher than guided search accuracy.

\begin{table}
	\caption{Count of words that are template and non-template.}
	\centering
	\begin{tabular}{ lccc }
		\toprule
		Count & TPT & Non-TPT & Total \\
		\midrule
		mean & 10.5 & 8.5 & 18.9 \\
min & 6 & 4 & 12 \\
max & 13 & 22 & 34 \\
		\bottomrule
	\end{tabular}
	\label{tab:count}
\end{table}

Table \ref{tab:incorrectcount} shows the summary statistics for the count of incorrect predictions made by the nvBERT model on the easy template words, hard template words and non-template words. The rate of incorrect prediction is much higher for hard template words than easy template words, as expected. The third column shows the statistics for incorrect prediction of non-template words, curiously the minimum of incorrect prediction is 1, which indicates that every prediction has at least one incorrectly predicted word, yet the guided search did result in completely correct predictions. The cause behind this apparent discrepancy is that for the guided search, predicted words are checked for whether it is possible for the word to occupy the position. Below shows an example:

\begin{table}
	\caption{Incorrect prediction for easy template, hard template, total template and non-template prediction count.}
	\centering
	\begin{tabular}{lcccc}
		\toprule
		Incorrect & Easy & Hard & TPT & Non-TPT \\
		\midrule
		mean & 0 & 0.6 & 0.6 & 3.4 \\
min & 0 & 0 & 0 & 1 \\
max & 1 & 4 & 4 & 12 \\
		\bottomrule
	\end{tabular}
	\label{tab:incorrectcount}
\end{table}

\begin{quote}
\gray{Query}: <N> List all the possible ways to get to attractions , together with the number of attractions accessible by these methods in a bar chart , and sort names in asc order please . </N> <C> mark [T] data tourist\_attractions encoding x [X] y aggregate [AggFunction] [Y] color [Z] transform filter [F] group [G] bin [B] sort [S] topk [K] </C> <D> tourist\_attractions <COL> name how\_to\_get\_there </COL> <VAL> walk bus </VAL> </D>
\end{quote}

\begin{quote}
\gray{Label}: mark bar data \red{tourist\_attractions} encoding x how\_to\_get\_there y aggregate count how\_to\_get\_there transform group x sort x asc <eos>
\end{quote}

\begin{quote}
\gray{Prediction}: mark bar data \red{employees} encoding x how\_to\_get\_there y aggregate count how\_to\_get\_there transform group x sort x asc <eos>
\end{quote}

Although the prediction has incorrectly predicted ``employees'' instead of ``tourist\_attractions'', ``employees'' is not the name of the table under consideration, therefore guided search chose the word that matches the table name instead, resulting in a correct prediction.

Table \ref{tab:accuracybreakdown} shows the breakdown of nvBERT model prediction accuracy by category. As expected, template words prediction accuracy is much higher than non-template words.

\begin{table}
	\caption{nvBERT prediction accuracy breakdown.}
	\centering
	\begin{tabular}{ccccc }
		\toprule
		Easy & Hard & TPT & Non-TPT & Total \\
		\midrule
		100\% & 89\% & 94\% & 60\% & 79\% \\
		\bottomrule
	\end{tabular}
	\label{tab:accuracybreakdown}
\end{table}

Finally, Table \ref{tab:chartaccuracy} shows the accuracy of the nvBERT model at predicting the chart type, ``arc'', ``bar'', ``line'' or ``point''. The accuracy is broken down by whether the chart type is already provided by the source query, interestingly, providing chart type in source query does not improve the accuracy of predicting chart type, which is also reflected in Table \ref{tab:accuracy}. However, the provided chart type does help guided search to determine whether the predicted chart type is valid or not, thereby improving the guided search accuracy \ref{tab:guidedaccuracy}.

\begin{table}
	\caption{nvBERT chart type prediction accuracy breakdown.}
	\centering
	\begin{tabular}{ccccc}
		\toprule
		Query & Query+Chart \\
		\midrule
		19\% & 19\% \\
		\bottomrule
	\end{tabular}
	\label{tab:chartaccuracy}
\end{table}

\subsubsection{CodeT5} For inference, the standard model decoder was used. Even though certain words such as table names and available column choices were encouraged via including them with the model input, no output tokens were forced. With this in mind the results from fine-tuning the CodeT5 model has been phenomenal, achieving a final token accuracy of 98\%. The model showed a good ability to predict the general structure of the query with a 96\% success rate predicting query keywords (i.e. “count”, “group”, “transform”, etc.) and in 93\% of the cases, was able to correctly identify the column names. A results review shows that in the majority of the cases, the error was semantic or abbreviation related. For example, for a label column name of `market\_val` the model would predict `market\_value`. In other cases the error was more related to the way the question was asked and the column name. For instance the model would predict “profit” instead of “profit\_dollars” because the question would ask for profits without mentioning dollars.

\begin{table}
	\caption{CodeT5 performance breakdown by difficulty level.}
	\centering
	\begin{tabular}{lccc}
		\toprule
		Hardness & Success & Failure & Success Rate \\
		\midrule
		Easy & 1451 & 147 & 91\% \\
		Medium & 1869 & 233 & 89\% \\
		Hard & 605 & 135 & 81\% \\
		Extra Hard & 384 & 52 & 88\% \\
		\bottomrule
	\end{tabular}
	\label{tab:codet5rate}
\end{table}

When comparing exact query matches as shown in Table \ref{tab:codet5rate}, the CodeT5 model consistently achieved high performance across difficulty levels, suggesting that the complexity of language and label query structure is not a significant limiting factor in its performance. A quick examination of the results show that in many of the cases the errors were systemic in nature and could easily be corrected by a series of logical filters. Some of these errors involved unmatched `\%` signs which are used for wildcards. For instance, to match values with the letter “a'' the pattern would be `\%a\%`. The other most common error involved a lack of a white-space preceding the unequal operator `!=`. Please refer to the appendix for examples. In fact, fixing this one error improved the query accuracy from 83\% to 88\% as can be seen in Table \ref{tab:codet5error}. The rest of the errors were more nuanced in nature and involved commonly being unable to correctly predict the transform sections which tended to be the most complex. In this area the CodeT5 model only achieved a 90\% success rate.

\begin{table}
	\caption{CodeT5 success rate before and after error correction.}
	\centering
	\begin{tabular}{cc}
		\toprule
		Before Correction & After Correction \\
		\midrule
		83\% & 88\% \\
		\bottomrule
	\end{tabular}
	\label{tab:codet5error}
\end{table}

\section{Conclusion}

Through our research and analysis, we have demonstrated the potential of transfer learning through the use of the BERT model as encoder for natural language to visualization tasks, specifically applied to the nvBench dataset. With the result of the CodeT5 model trained and tested on the nvBench dataset we see that the problem could be adequately modeled as a natural language to code translation problem. With more sophisticated architecture and more comprehensive data visualization language, it is clear that natural language to data visualization translation could be done very effectively and in a general manner with existing large language models.

Of course, there are still many areas that remain to be explored. First, a more generic visualization language needs to be developed to accommodate more complex requests, the design of this language should be guided by the effectiveness with which it could be predicted by language models. Second, more database design should take into account the need for visualization, and integrate machine learning translation as part of the research and development process. We hope our research would contribute to the progress of these initiatives.

%----------------------------------------------------------------------------------------
%	 REFERENCES
%----------------------------------------------------------------------------------------

\phantomsection
\bibliographystyle{unsrt}
\bibliography{t0003.bib}

\begin{thebibliography}{1}

\bibitem{Zhong:2017qr}
Victor. Zhong et~al.
\newblock Seq2sql: Generating structured queries from natural language using
  reinforcement learning.
\newblock {\em arXiv:1709.00103v7 [cs.CL]}, November 2017.

\bibitem{Luo:2021qr}
Yuyu. Luo et~al.
\newblock nvbench: A large-scale synthesized dataset for cross-domain natural
  language to visualization task.
\newblock {\em arXiv:2112.12926v1 [cs.HC]}, December 2021.

\bibitem{Yu:2019qr}
Tao. Yu et~al.
\newblock Cosql: A conversational text-to-sql challenge towards cross-domain
  natural language interfaces to databases.
\newblock {\em arXiv:1909.05378v1 [cs.CL]}, September 2019.

\bibitem{Cox:2001qr}
Kenneth. Cox et~al.
\newblock A multi-modal natural language interface to an information
  visualization environment.
\newblock {\em International Journal of Speech Technology}, 2001.

\bibitem{Luo:2022qr}
Yuyu. Luo et~al.
\newblock Natural language to visualization by neural machine translation.
\newblock {\em IEEE Transactions on Visualization and Computer Graphics, vol.
  28, no. 1, pp. 217-226}, January 2022.

\bibitem{Aurisano:2016qr}
Jillian. Aurisano et~al.
\newblock Articulate2: Toward a conversational interface for visual data
  exploration.
\newblock {\em IEEE VIS ’16 (Poster paper)}, January 2016.

\bibitem{Can:2021qr}
Can. Liu et~al.
\newblock Advisor: Automatic visualization answer for natural-language question
  on tabular data.
\newblock {\em 2021 IEEE 14th Pacific Visualization Symposium (PacificVis)},
  January 2021.

\bibitem{Raffel:2020qr}
Colin. Raffel et~al.
\newblock Exploring the limits of transfer learning with a unified text-to-text
  transformer.
\newblock {\em Journal of Machine Learning Research}, 2020.

\bibitem{Yue:2021qr2}
Yue. Wang et~al.
\newblock Codet5: Identifier-aware unified pre-trained encoder-decoder models
  for code understanding and generation.
\newblock {\em Conference on Empirical Methods in Natural Language Processing},
  2021.

\end{thebibliography}

%----------------------------------------------------------------------------------------

\clearpage
\begin{appendices}

\section*{APPENDIX}

\subsection{Model Architectures}

\begin{figure}[H]
	\includegraphics[width=\linewidth]{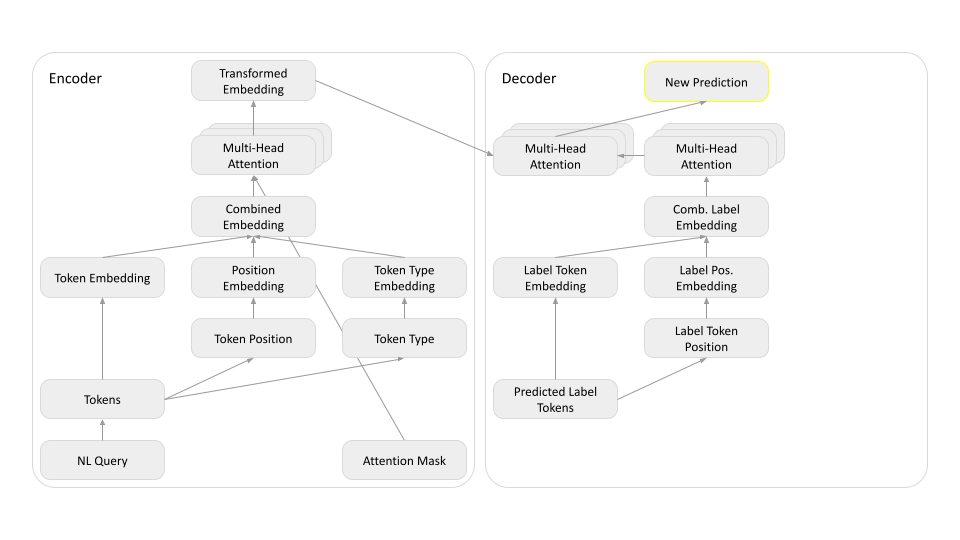}
	\caption{Architecture of ncNet model.}
	\label{fig:ncnet}
\end{figure}

\begin{figure}[H]
	\includegraphics[width=\linewidth]{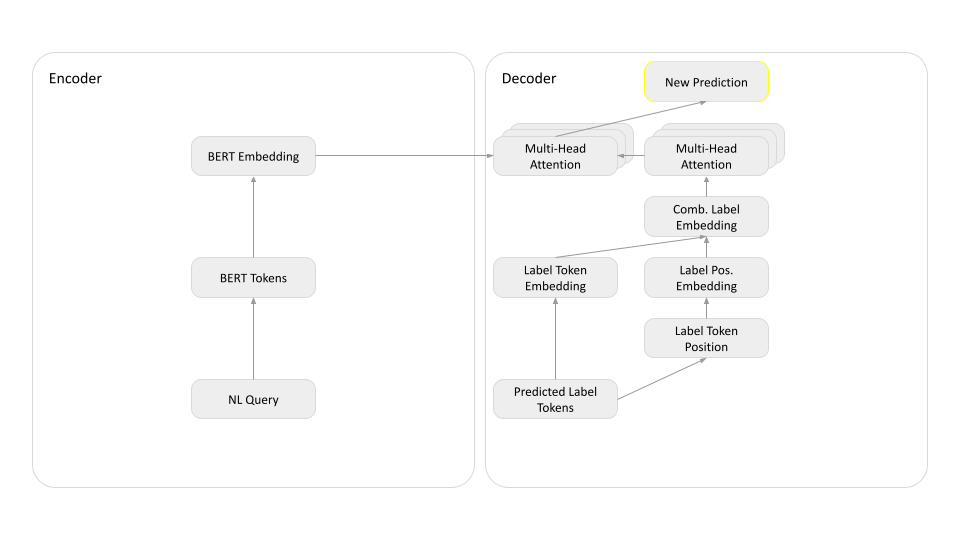}
	\caption{Architecture of nvBERT model.}
	\label{fig:ncbert}
\end{figure}

\begin{figure}[H]
	\includegraphics[width=\linewidth]{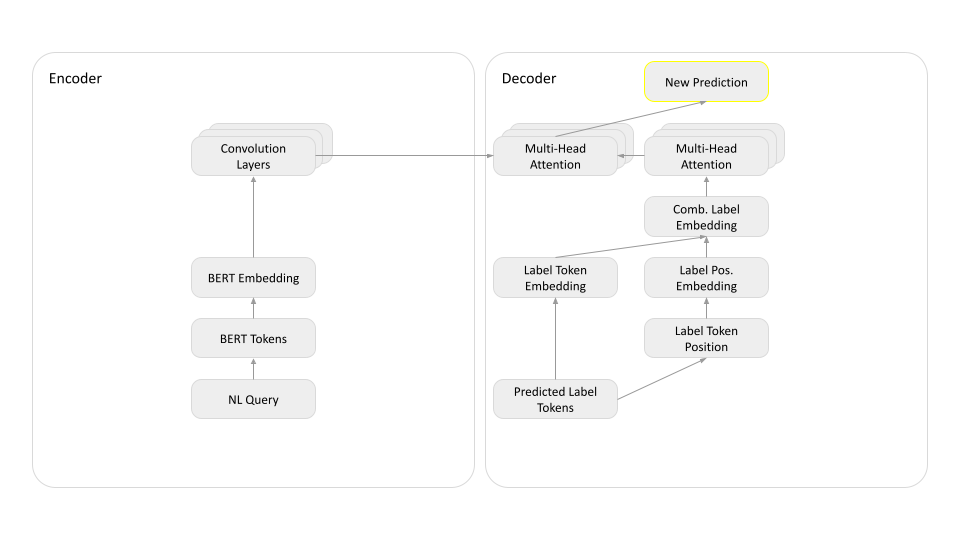}
	\caption{Architecture of nvBERT\_CNN model.}
	\label{fig:ncbertcnn}
\end{figure}

\begin{figure}[H]
	\includegraphics[width=\linewidth]{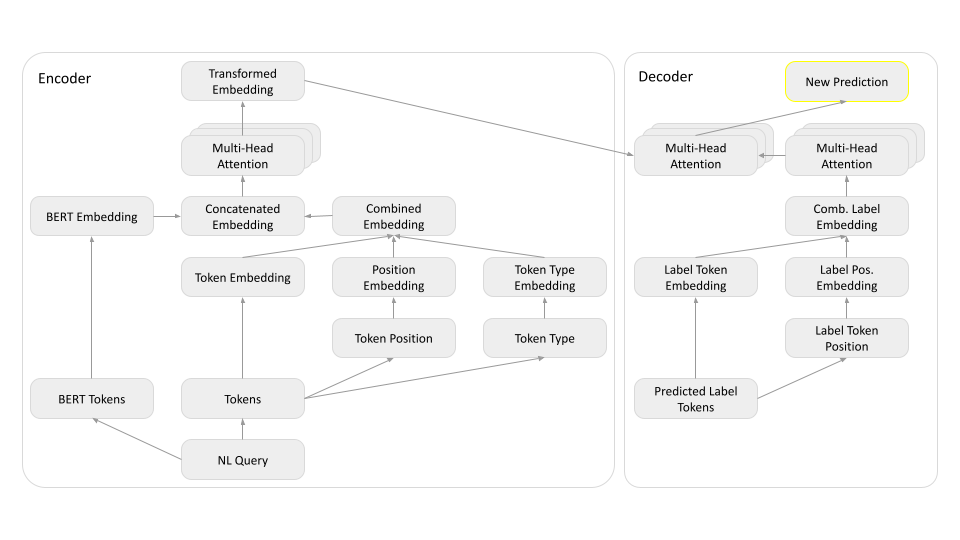}
	\caption{Architecture of nvncNetBERT model.}
	\label{fig:ncnetbert}
\end{figure}

\begin{figure}[H]
	\includegraphics[width=\linewidth]{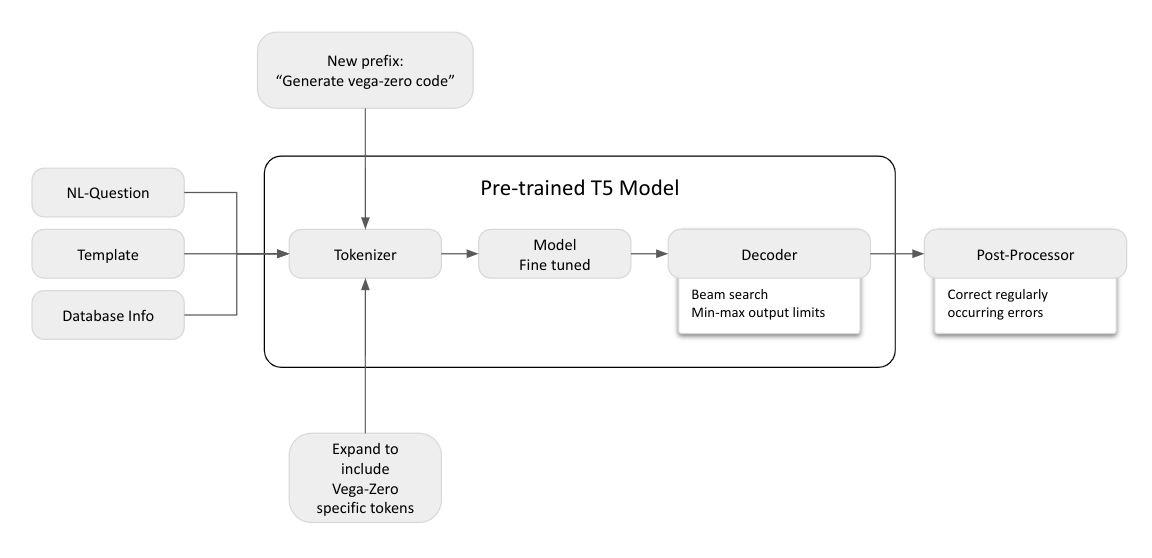}
	\caption{Architecture of CodeT5 models applied to nvBench.}
	\label{fig:codetf}
\end{figure}

\subsection{CodeT5 Error Examples}

We further examine some of the cases where the codeT5 model fails to predict the label exactly:

\begin{quote}
\gray{Label}: mark bar data customer encoding x cust\_name y aggregate none acc\_bal transform filter cust\_name like \red{'\%a\%'} sort y desc
\end{quote}

\begin{quote}
\gray{Prediction}: mark bar data customer encoding x cust\_name y aggregate none acc\_bal transform filter cust\_name like \red{'\%a'} sort y desc
\end{quote}

In the first example, the predicted vega zero query is missing a ``\%'' sign after ``\%a''.

\begin{quote}
\gray{Label}: mark bar data employees encoding x hire\_date y aggregate count hire\_date transform filter salary between 8000 and 12000 and commission\_pct != "null" or \red{department\_id != 40} sort y asc bin x by 'weekday
\end{quote}

\begin{quote}
\gray{Prediction}: mark bar data employees encoding x hire\_date y aggregate count hire\_date transform filter salary between 8000 and 12000 and commission\_pct!= "null" or \red{department\_id!= 40} sort y asc bin x by weekday
\end{quote}

In the second example, a space is missing before ``!=''. We have systematically corrected these mistakes for our final success rate. Table \ref{tab:codet5error} shows the improvement of results.

\subsection{Training and Loss}

\begin{figure*}
	\includegraphics[width=\linewidth]{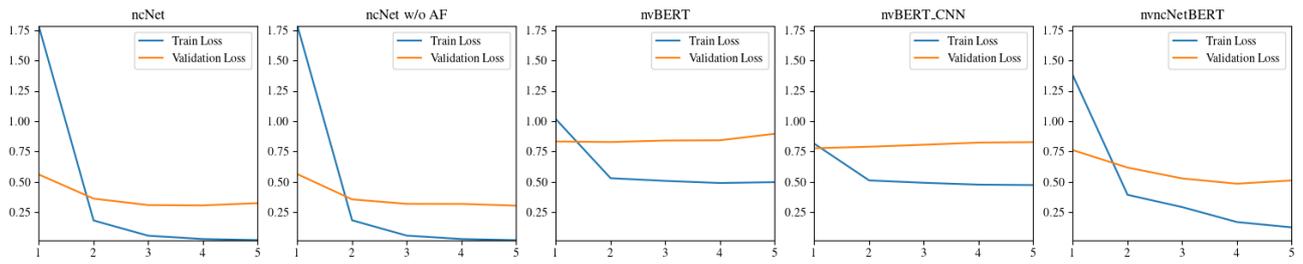}
	\caption{Train and validation loss of models.}
	\label{fig:loss}
\end{figure*}

The results of train and validation loss for ncNet, ncNet without attention forcing, nvBERT, nvBERT\_CNN and nvncNetBERT are shown for the first 5 epochs in Figure \ref{fig:loss}. The BERT encoder models show much smaller drop in train loss from epoch to epoch, due to transfer learning that occurs. However the BERT encoder models do not show significant drop in validation loss, suggesting that most of the fine-tuning has happened in the first epoch.

\end{appendices}
\end{document}